\documentclass[]{fairmeta}

\usepackage{enumitem}
\usepackage{booktabs}
\usepackage{subcaption}
\usepackage{float}

\title{VoxelCodeBench: Benchmarking 3D World Modeling Through Code Generation}

\author[1,2,*]{Yan Zheng}
\author[1]{Florian Bordes}

\affiliation[1]{FAIR at Meta}
\affiliation[2]{The University of Texas at Austin}

\contribution[*]{Work done at Meta}

\abstract{
Evaluating code generation models for 3D spatial reasoning requires executing generated code in realistic environments and assessing outputs beyond surface-level correctness. We introduce a platform VoxelCode, for analyzing code generation capabilities for 3D understanding and environment creation. Our platform integrates natural language task specification, API-driven code execution in Unreal Engine, and a unified evaluation pipeline supporting both automated metrics and human assessment. To demonstrate its utility, we construct VoxelCodeBench, a benchmark of voxel manipulation tasks spanning three reasoning dimensions: symbolic interpretation, geometric construction, and artistic composition. Evaluating leading code generation models, we find that producing executable code is far easier than producing spatially correct outputs, with geometric construction and multi-object composition proving particularly challenging. By open-sourcing our platform and benchmark, we provide the community with extensible infrastructure for developing new 3D code generation benchmarks and probing spatial reasoning in future models.}

\date{\today}
\correspondence{Florian Bordes at \email{fbordes@meta.com}}

\metadata[Code]{\url{https://github.com/facebookresearch/voxelcodebench}}

\begin{document}

\maketitle

\section{Introduction}
\label{sec:intro}

\begin{figure*}[t]
\centering
\includegraphics[width=\textwidth]{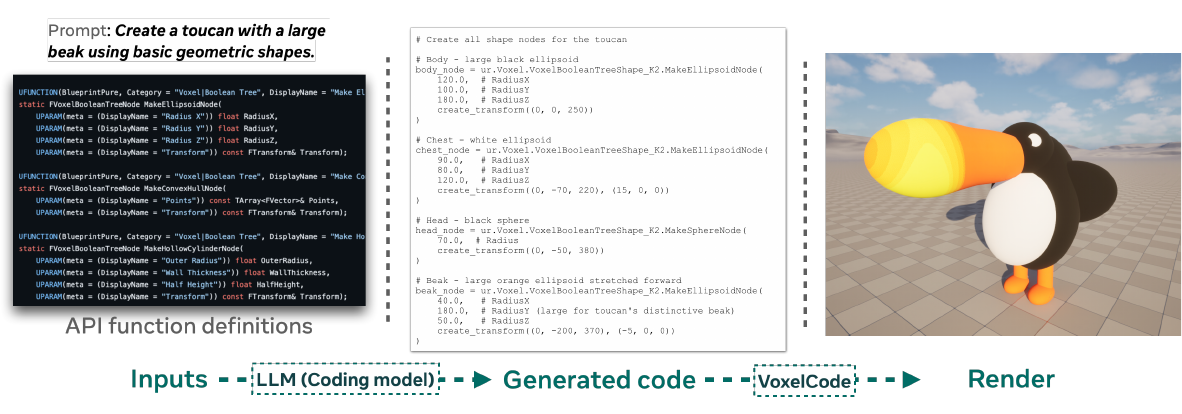}
\caption{\textbf{VoxelCodeBench evaluation pipeline.} Given API documentation and a natural language task specification (left), models generate Python code manipulating voxels in 3D space (center). The generated code is executed within VoxelCode, an Unreal Engine based rendering server equipped with the Voxel Plugin 2.0\citep{voxelplugin2}, producing visual outputs that can be evaluated by human annotators (right). Our benchmark spans three complexity tiers—from basic geometric primitives to compositional artistic scenes—enabling systematic assessment of spatial reasoning through code generation.}
\label{fig:teaser}
\end{figure*}

Code generation has emerged as a compelling paradigm for measuring and advancing language model capabilities~\citep{chen2021evaluating,li2022competition}. While existing benchmarks predominantly evaluate algorithmic problem-solving~\citep{austin2021program} or general programming tasks~\citep{zheng2023codegeex}, they do not assess spatial reasoning—the ability to understand, manipulate, and compose 3D geometric structures. This gap is particularly significant as spatial intelligence underlies critical applications including robotics~\citep{tellex2011understanding}, embodied AI~\citep{savva2019habitat}, and interactive content creation~\citep{wu2023omniobject3d}.

We argue that \emph{code generation for API-driven 3D manipulation} provides a principled framework for evaluating spatial reasoning in language models. Unlike end-to-end neural approaches~\citep{poole2022dreamfusion,lin2023magic3d}, such as text-to-3D generative models that work as black boxes, code-based methods expose the reasoning process through explicit symbolic manipulation, enabling fine-grained analysis of geometric understanding, compositional thinking, and creative problem-solving. Moreover, API-based evaluation reflects realistic software engineering scenarios where models must learn and apply domain-specific interfaces from documentation and examples. Lastly, in addition to better analysis, leveraging code as an intermediate step for 3D generation allows users to have more control over the output.

In this work, we introduce VoxelCode, a rendering API, built with Unreal Engine\citep{unreal_disclaimer} and the Voxel Plugin 2.0\citep{voxelplugin2}, that takes code as input and produces real-time 3D rendering as output. This platform allows users to instruct a Large Language Model (LLM) to generate 3D objects using code. A key feature is the ability to manipulate a virtual camera, enabling the generation of multiple screenshots from various viewpoints. This functionality makes VoxelCode an excellent tool for directly comparing the coding abilities of different LLMs by visualizing and assessing their performance in terms of visual consistency and quality when working with simple 3D primitives.

Using VoxelCode, we developed \textsc{VoxelCodeBench}, a comprehensive benchmark for systematically evaluating code generation models on 3D voxel manipulation tasks. Our benchmark addresses three core research questions: \textbf{(RQ1)} Can contemporary code generation models learn to use specialized 3D manipulation APIs from documentation and examples? \textbf{(RQ2)} How do model capabilities scale across symbolic, geometric, and artistic reasoning dimensions? \textbf{(RQ3)} What are the relative strengths of general-purpose LLMs versus code-specialized models for spatial reasoning tasks?

\textsc{VoxelCodeBench} comprises 220 carefully designed tasks organized along three axes of increasing complexity:
\begin{itemize}[leftmargin=*,noitemsep,topsep=2pt]
\item \textbf{Symbolic reasoning} (80 tasks): Coordinate-based primitive placement, testing fundamental API comprehension and spatial coordinate mapping
\item \textbf{Geometric construction} (50 tasks): Procedural shape generation with boolean operations, evaluating compositional geometric thinking
\item \textbf{Artistic composition} (90 tasks): Multi-object scene creation with aesthetic constraints, measuring creative spatial reasoning and thematic coherence
\end{itemize}

\paragraph{Key contributions.}
\begin{enumerate}[leftmargin=*,noitemsep,topsep=2pt]
\item We introduce \textsc{VoxelCode}, an unreal-engine based platform that render 3D objects from simple python code using an API that leverage only basic geometries.
\item We introduce \textsc{VoxelCodeBench}, a systematic benchmark for evaluating 3D spatial reasoning through API-driven code generation, comprising 200+ tasks with standardized evaluation protocols and comprehensive human annotations.
\item Through extensive experiments across models, we establish quantitative baselines and identify systematic performance patterns across spatial reasoning dimensions.
\item We release our complete benchmark infrastructure—including task dataset, rendering environment, and evaluation tools—to facilitate future research in code-based spatial reasoning.
\end{enumerate}

Our findings reveal that while contemporary models demonstrate robust API learning capabilities, significant gaps remain in compositional geometric reasoning and creative spatial composition. These insights have implications for both improving code generation models and designing effective human-AI collaboration systems for 3D content creation.

\section{Related Work}
\label{sec:related}

\paragraph{Benchmarks for code generation.}
Recent benchmarks for evaluating code generation models primarily focus on algorithmic problem-solving and general programming tasks. HumanEval~\citep{chen2021evaluating} and MBPP~\citep{austin2021program} assess function-level code generation from natural language descriptions, while CodeContests~\citep{li2022competition} targets competitive programming problems. APPS~\citep{hendrycks2021measuring} introduces more complex programming challenges but still emphasizes algorithmic reasoning over domain-specific API usage. More recent work has explored domain-specific code generation for data science~\citep{lai2023ds1000} and API usage~\citep{liu2023apibank}, but these benchmarks do not assess spatial reasoning or 3D manipulation capabilities. Other works, such as SceneCraft \citep{scenecraft}, utilize LLMs to synthesize novel scenes by generating Blender code. Similarly, \citep{du2024blenderllmtraininglargelanguage} also focuses on computer-aided design using language models.

\paragraph{3D generation and evaluation.}
Text-to-3D generation has seen rapid progress through neural rendering approaches including NeRF optimization~\citep{jain2022zero}, diffusion-based methods~\citep{poole2022dreamfusion,lin2023magic3d,wang2023prolificdreamer}, and 3D Gaussian splatting~\citep{tang2023dreamgaussian}. While these methods produce high-quality visual outputs, evaluation typically focuses on single-object generation quality~\citep{wu2023omniobject3d,deitke2023objaverse} rather than compositional reasoning or procedural construction capabilities. Recent benchmarks like T3Bench~\citep{he2023t3bench} evaluate compositional text-to-3D generation but through end-to-end neural systems rather than explicit code generation. Our benchmark complements this line of work by providing a code-based evaluation framework that enables analysis of compositional spatial reasoning through interpretable symbolic manipulation.

\paragraph{CSG-based shape program synthesis.}
Recent work explores inferring programmatic representations from 3D shapes. MeshCoder~\citep{dai2025meshcoder} trains a
multimodal LLM to translate point clouds into Blender scripts, while PrimitiveAnything~\citep{ye2025primitiveanything}
learns primitive decomposition from human-annotated assemblies. However, these methods solve the \emph{inverse} problem
(shape$\rightarrow$code) and require category-specific training data, limiting generalization. In contrast,
\textsc{VoxelCodeBench} evaluates the \emph{forward} problem (language$\rightarrow$code$\rightarrow$shape) in a zero-shot
setting, testing whether pretrained LLMs can generalize to novel 3D tasks without domain-specific training.

\paragraph{Spatial reasoning in language models.}
Prior work has explored spatial reasoning capabilities in language models through various modalities. Early work focused on textual spatial descriptions~\citep{mirzaee2021spartqa,liu2023visual}, while recent efforts incorporate visual grounding~\citep{zhang2023gpt4roi,yang2023dawn}. Embodied AI benchmarks like ALFRED~\citep{shridhar2020alfred} and Habitat~\citep{savva2019habitat} evaluate spatial reasoning through interactive navigation and manipulation but do not assess code generation capabilities. Most relevant to our work, recent studies examine LLM spatial reasoning through code generation for robotics~\citep{liang2023code,singh2023progprompt} and interactive environments~\citep{wang2023voyager}.

\paragraph{Voxel-based representations.}
Voxel representations provide discrete, regular grids for 3D data and have been widely adopted in computer vision~\citep{maturana2015voxnet,riegler2017octnet,choy20163d} and graphics~\citep{laine2010efficient}. Recent work explores learning-based voxel generation through GANs~\citep{wu20153d} and diffusion models~\citep{hui20223d}. In the context of LLM applications, voxel-based environments like Minecraft have been used for studying embodied agents~\citep{fan2022minedojo,wang2023voyager} but without systematic benchmarking of code generation for spatial reasoning. Our work leverages voxel representations as an intuitive substrate for evaluating API-driven spatial manipulation, providing discrete building blocks amenable to programmatic construction while supporting complex geometric operations through boolean algebra.

\section{VoxelCode}

\begin{figure}[t]
\centering
\includegraphics[width=0.6\textwidth]{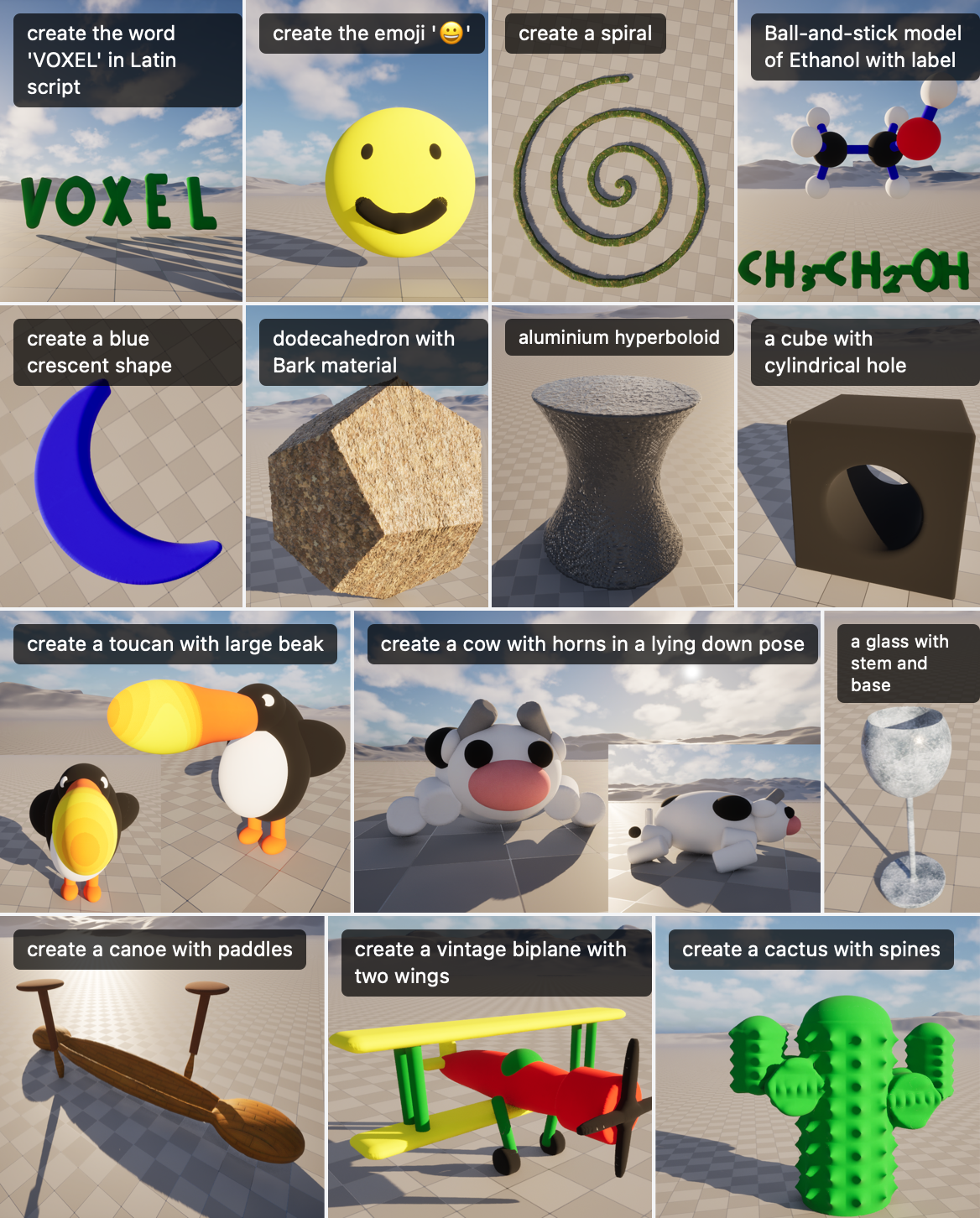}
\caption{\textbf{Example outputs using VoxelCode.} Representative 3D voxel constructions generated by models across our benchmark, including characters and symbols, geometric and mathematical shapes, artistic animals, objects, vehicles, and natural objects, ranging from simple geometric primitives to complex multi-component objects.}
\label{fig:examples}
\end{figure}

Our first contribution is VoxelCode, an environment built with Unreal Engine that takes Python code as input and executes it to manipulate voxels. Figure~\ref{fig:teaser} illustrates the overall pipeline: given API documentation and a natural language prompt, a model generates Python code that VoxelCode executes to produce rendered 3D scenes. To build VoxelCode, we leverage VoxelPlugin 2.0~\citep{voxelplugin2}, which allows for the precise manipulation of voxels within Unreal Engine environments. In contrast to many traditional 3D rendering software, such as the easily accessible PyVista~\citep{sullivan2019pyvista} which comes with extensive examples and documentations, the Voxel Plugin 2.0 source code is not publicly available and requires a custom license. Using a tool that is not widely accessible is crucial for ensuring that large language models (LLMs) are indeed generalizing and not over-fitting on their training data. The Voxel Plugin is compiled within VoxelCode, making its source code inaccessible. The only way to communicate with the Voxel Plugin when using VoxelCode is by leveraging a simple API we made, in which we only present some header function definitions and python examples scripts that a model can leverage. In addition to licensing Unreal Engine and the Voxel Plugin 2.0, we also licensed materials data that a model can use to improve its rendering. VoxelCode runs on Linux servers using the Vulkan API and professional gaming GPUs such as the Nvidia L40 or Nvidia RTX 6000. Alongside VoxelCode, we provide a web-interface that allows users to write code and visualize the code execution in real-time through WebRTC. Figure~\ref{fig:examples} shows representative outputs spanning characters, geometric shapes, animals, vehicles, and architectural structures—ranging from simple primitives to complex multi-component objects.

\section{VoxelCodeBench Design}
\label{sec:method}

We design \textsc{VoxelCodeBench} to enable systematic, reproducible evaluation of spatial reasoning through code generation. Our benchmark architecture consists of three core components: (1) a task specification framework providing API documentation and natural language prompts, (2) a VoxelCode execution infrastructure, and (3) multi-dimensional assessment protocols. Figure~\ref{fig:teaser} illustrates this pipeline end-to-end.

\subsection{Evaluation Framework}

Each task in \textsc{VoxelCodeBench} provides models with three components: (1) minimal API documentation, (2) minimal working examples, and (3) natural language task specification.

\paragraph{API documentation.} We provide documentation for a voxel manipulation API comprising 47 functions organized into four categories: \emph{primitive placement} (individual voxel operations), \emph{geometric shapes} (parameterized 3D primitives), \emph{boolean operations} (set-theoretic combinations), and \emph{surface utilities} (mesh generation and smoothing). Each function includes type signatures, parameter descriptions, coordinate system specifications, and constraint documentation. This mirrors realistic software engineering scenarios where models must learn domain-specific APIs from available documentation.

\paragraph{Working examples.} For each task category, we provide 2-3 minimal working examples demonstrating correct API usage patterns, parameter specification, and common idioms. These examples use the same functions required for task completion but solve simpler problems, testing models' ability to generalize from demonstrations.

\paragraph{Task specifications.} Natural language descriptions specify desired 3D entities with varying levels of detail and constraint. Specifications range from precise geometric descriptions (\textit{"Create a red cube with side length 5 centered at origin"}) to open-ended creative prompts (\textit{"Build a medieval castle with towers and defensive walls"}). This variation enables assessment across the spectrum from instruction-following to creative problem-solving.

\subsection{Task Taxonomy}
\textsc{VoxelCodeBench} organizes 220 tasks along three axes representing distinct facets of spatial reasoning:

\paragraph{Symbolic reasoning (80 tasks).} These tasks evaluate fundamental API comprehension and coordinate mapping capabilities. Subtasks include: (i) \emph{discrete placement} (placing individual voxels at specified coordinates), (ii) \emph{pattern reconstruction} (recreating 2D patterns and character shapes in 3D space), (iii) \emph{primitive instantiation} (generating basic shapes with specified parameters), (iv) \emph{simple arrangements} (spatial configurations of multiple primitives), (v) \emph{molecular models} (ball-and-stick representations of chemical compounds), and (vi) \emph{writing systems} (rendering text, characters, and symbols from various scripts). Success on symbolic tasks indicates models can parse API documentation, map natural language specifications to function calls, and perform basic coordinate calculations.

\paragraph{Geometric construction (50 tasks).} These tasks assess procedural shape generation and compositional reasoning. Subtasks include: (i) \emph{parameterized primitives} (creating complex geometric forms like arches, spirals, helices), (ii) \emph{boolean composition} (combining shapes through set operations), (iii) \emph{iterative construction} (building structures through loops and recursion), and (iv) \emph{surface generation} (creating smooth meshes from voxel data). Success requires understanding geometric principles, spatial composition, and algorithmic construction strategies.

\paragraph{Artistic composition (90 tasks).} These tasks measure creative spatial reasoning and aesthetic judgment. Subtasks include: (i) \emph{thematic scenes} (multi-object environments with stylistic coherence), (ii) \emph{functional structures} (buildings, vehicles, objects with purpose), (iii) \emph{natural forms} (organic shapes like trees, animals), and (iv) \emph{abstract compositions} (artistic arrangements emphasizing visual balance). Success demands integration of geometric skills with creative problem-solving and world knowledge about object appearance and spatial relationships.

\subsection{VoxelCode Environment}

Generated code is executed through the VoxelCode library. The execution pipeline: (1) validates code syntax and imports, (2) initializes clean scene state, (3) executes generated code with 60-second timeout, (4) captures execution logs and error messages, (5) renders final scene from multiple viewpoints (front, side, top, perspective), and (6) exports geometric data for automated analysis.

This infrastructure ensures reproducible evaluation across models while providing rich feedback for error analysis.

\subsection{Multi-Dimensional Assessment}
To comprehensively evaluate model performance, \textsc{VoxelCodeBench} combines automated metrics with human evaluation, providing both objective correctness measures and subjective quality assessments.

\paragraph{Automated metrics.}
We first apply two automated checks to filter obvious failures. The \textbf{execution success} metric verifies that generated code runs without syntax or runtime errors, while \textbf{API compliance} detection identifies common violations such as invalid parameters and constraint violations.

\paragraph{Human evaluation.} 
While automated metrics catch clear errors, they cannot assess whether the generated 3D scene actually matches the intended specification. We therefore conduct systematic human evaluation using a custom annotation interface (Figure~\ref{fig:humanAnno}). For each generated output, annotators evaluate five key aspects: whether any visible 3D object is present (\textbf{Has Object}), whether objects are accurately placed at specified coordinates (\textbf{Position Correctness}), whether specified materials and colors are correctly applied (\textbf{Material Correctness}), whether the generated geometry aligns with the specification (\textbf{Shape Correctness}), and the overall aesthetic quality on a 0--10 scale (\textbf{Visual Quality}). This design allows us to disentangle different aspects of the generation task—distinguishing, for instance, between models that create the right shape in the wrong location versus those that fail at shape generation entirely.

\paragraph{Task completion criterion.} 
We treat \textbf{Shape Correctness} as our primary success metric, since it most directly captures whether models can accurately translate language into 3D geometry. We use \textbf{Visual Quality} as a secondary metric to assess aesthetic aspects beyond binary correctness.

\section{Experiments and results}
\label{sec:experiments}
\label{sec:results}

We evaluate eight state-of-the-art large language models across three major families: three models from OpenAI's GPT family (GPT-5, GPT-5 Mini, and GPT-5 Chat), four from Anthropic's Claude series (Claude Sonnet 4.5, Claude Opus 4, Claude 3.5 Sonnet, and Claude 3 Opus), and Gemini 3 Pro from Google. This selection allows us to compare both across and within model families, examining how different capability tiers and training approaches affect performance on spatial reasoning tasks.

We access all models through their respective APIs under identical evaluation conditions. Each model receives the same input consisting of complete API documentation, working code examples, and the natural language task specification, ensuring that performance differences reflect genuine capability gaps rather than variations in available context.

\subsection{Evaluation Protocol}

\begin{figure}[t]
\centering
\includegraphics[width=0.6\textwidth]{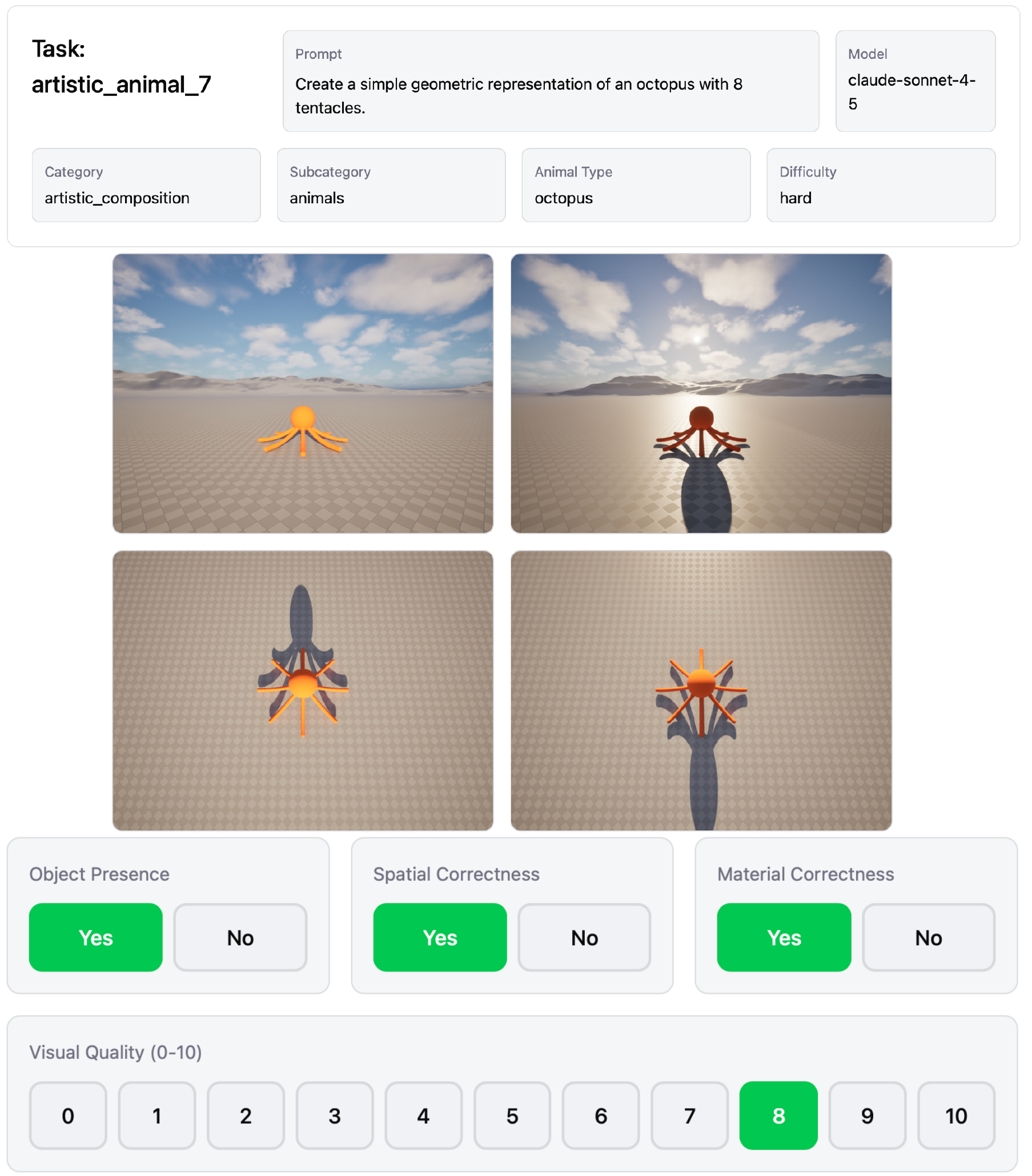}
\caption{\textbf{Human annotation interface.} Annotators view the prompt and rendered outputs from multiple viewpoints,
then rate five dimensions: Has Object, Position Correct, Material Correct, Shape Correct, and Visual Quality (0--10).}
\label{fig:humanAnno}
\end{figure}

\paragraph{Human annotation.} We developed a custom annotation interface (Figure~\ref{fig:humanAnno}) to evaluate model outputs across five dimensions:
Has Object, Position Correct, Material Correct, Shape Correct, and Visual Quality (0--10). The interface displays task
metadata, the original prompt, and rendered outputs from four viewpoints, enabling systematic assessment of each
generation.

\paragraph{Metrics.} Our primary metric is \textbf{Shape Correct}: the percentage of tasks where the generated shape
matches the specification. We additionally report Position Correct, Material Correct, and Visual Quality scores to provide
fine-grained analysis of model capabilities.

For each task, generated code is executed in VoxelCode, which renders the scene from 4 viewpoints at 1920$\times$1080
resolution. The rendered images are then passed to our annotation interface for human evaluation.

\subsection{Results}
\label{sec:main_results}

\begin{figure*}[t]
\centering
\includegraphics[width=0.95\textwidth]{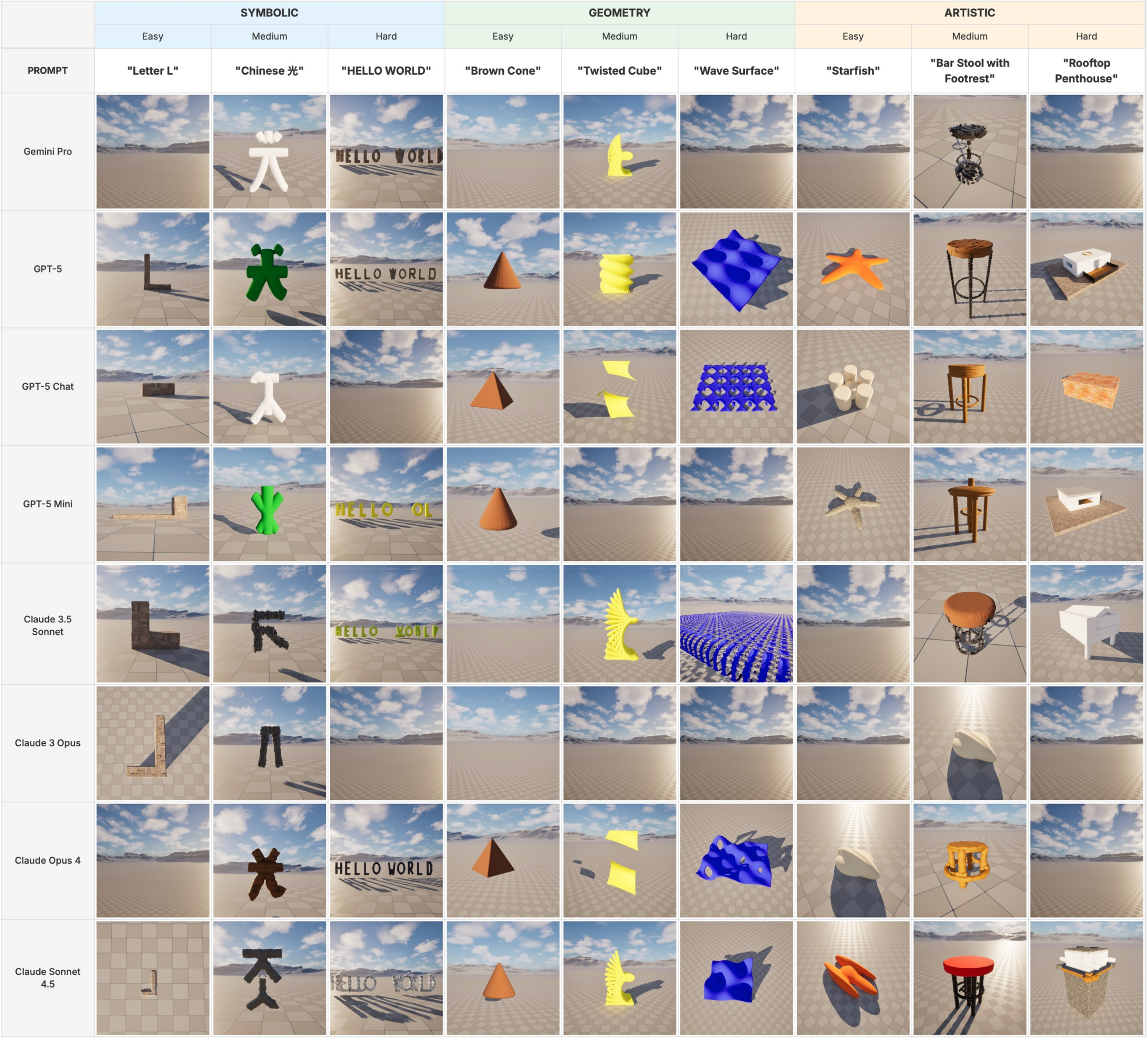}
\vspace{-6pt}
\caption{\textbf{Qualitative comparison across models, categories, and difficulty levels.} Rows correspond to eight evaluated models; columns span three task categories (Symbolic, Geometric, Artistic) at Easy, Medium, and Hard difficulty. Empty cells indicate cases where code execution failed, resulting in no visible output due to errors in the model-generated code.}
\vspace{-6pt}
\label{fig:qualitative}
\end{figure*}

Table~\ref{tab:main_results} summarizes model performance across \textsc{VoxelCodeBench}, evaluating whether outputs contain visible objects (\textbf{HasObj}), the correctness of position, material, and shape attributes, and overall visual quality (\textbf{Look}, scored 0--10). Figure~\ref{fig:qualitative} provides qualitative examples spanning all models, task categories, and difficulty levels.

\begin{table}[t]
\centering
\caption{\textbf{Main results on VoxelCodeBench.} Performance metrics (\%) across models. \textbf{HasObj}: output contains visible object. \textbf{Pos/Mat/Shape}: position, material, shape correctness. \textbf{Look}: average visual quality score (0--10). Best results in \textbf{bold}.}
\label{tab:main_results}
\resizebox{0.55\textwidth}{!}{%
\begin{tabular}{lccccc}
\toprule
\textbf{Model} & \textbf{HasObj} & \textbf{Pos} & \textbf{Mat} & \textbf{Shape} & \textbf{Look} \\
\midrule
GPT-5 & \textbf{88.6} & \textbf{88.6} & \textbf{87.9} & \textbf{87.9} & \textbf{5.71} \\
GPT-5 Mini & 85.8 & 85.8 & 82.4 & 80.4 & 4.86 \\
Claude Sonnet 4.5 & 86.7 & 86.7 & 84.6 & 80.4 & 5.01 \\
GPT-5 Chat & 77.9 & 77.9 & 75.9 & 69.7 & 3.66 \\
Claude Opus 4 & 74.1 & 72.8 & 72.1 & 69.4 & 4.13 \\
Claude 3.5 Sonnet & 71.0 & 70.3 & 70.3 & 66.9 & 3.30 \\
Claude 3 Opus & 59.5 & 59.5 & 54.8 & 45.2 & 3.40 \\
Gemini Pro & 20.1 & 20.1 & 18.8 & 19.5 & 1.36 \\
\bottomrule
\end{tabular}}
\end{table}

GPT-5 achieves the strongest overall performance, reaching 87.9\% shape correctness with the highest visual quality score of 5.71 out of 10. Claude Sonnet 4.5 and GPT-5 Mini follow closely, both achieving 80.4\% shape correctness, though GPT-5 Mini slightly edges ahead on visual quality. The results reveal interesting patterns within model families. The GPT series demonstrates relatively consistent performance across capability tiers, with even the smallest variant (GPT-5 Mini) maintaining strong results. The Claude family exhibits more substantial variation—Claude Sonnet 4.5 performs comparably to top GPT models at 80.4\%, while Claude 3 Opus lags significantly behind at just 45.2\% shape correctness, a 35-point gap within the same family.

Gemini Pro presents a stark outlier, achieving only 19.5\% shape correctness and a visual quality score of 1.36. This 68-point gap compared to GPT-5 suggests fundamental difficulties with the voxel manipulation API that go beyond mere performance degradation. The model appears to struggle with basic API comprehension, potentially failing to properly parse the documentation or generalize from the provided examples.

\begin{table}[t]
\centering
\caption{\textbf{Performance by task category.} Shape correctness (\%) and visual quality (Look, 0--10) across three reasoning dimensions. Best per category in \textbf{bold}.}
\label{tab:category_results}
\resizebox{0.7\textwidth}{!}{%
\begin{tabular}{l|cc|cc|cc}
\toprule
& \multicolumn{2}{c|}{\textbf{Symbolic}} & \multicolumn{2}{c|}{\textbf{Geometric}} & \multicolumn{2}{c}{\textbf{Artistic}} \\
\textbf{Model} & Shape & Look & Shape & Look & Shape & Look \\
\midrule
GPT-5 & 87.5 & \textbf{6.81} & \textbf{66.7} & \textbf{6.56} & \textbf{97.5} & \textbf{4.90} \\
Claude Sonnet 4.5 & \textbf{90.3} & 6.77 & 52.8 & 5.28 & 89.5 & 4.16 \\
Claude Opus 4 & 75.0 & 6.25 & 40.0 & 3.83 & 80.0 & 3.41 \\
Gemini Pro & 15.2 & 1.55 & 23.7 & 2.08 & 19.2 & 0.92 \\
\bottomrule
\end{tabular}}
\end{table}

Breaking down performance by task category (Table~\ref{tab:category_results}) reveals that different models excel at different types of spatial reasoning. For symbolic tasks requiring precise API usage and spatial coordinates, Claude Sonnet 4.5 slightly outperforms GPT-5 (90.3\% vs. 87.5\%), though both achieve visual quality scores above 6.8. Gemini Pro's collapse to 15.2\% on these tasks reinforces the API comprehension hypothesis—symbolic reasoning requires faithful adherence to documentation, where Gemini Pro consistently fails.

Geometric construction emerges as the most challenging category across all models. Even GPT-5, the overall leader, drops to 66.7\% shape correctness—a 21-point decline from its symbolic performance. Claude Sonnet 4.5 falls more sharply to 52.8\%. Digging into the subcategories reveals an intriguing divergence in spatial reasoning strategies. GPT-5 excels at geometry operations like unions and subtractions (84.2\%) but struggles with parameterized primitives such as spheres and cylinders (54.5\%). Claude Sonnet 4.5 exhibits precisely the opposite pattern, achieving 54.5\% on parameterized primitives but only 42.1\% on geometry operations. This suggests the models have developed fundamentally different internal representations of 3D space—GPT-5 appears stronger at relational reasoning between objects, while Claude Sonnet 4.5 handles individual object parameterization more reliably.

Surprisingly, artistic composition tasks yield the strongest results for top-tier models. GPT-5 achieves 97.5\% shape correctness, reaching perfect 100\% success on both animals and everyday objects. Claude Sonnet 4.5 maintains strong performance at 89.5\%. This pattern likely reflects the interpretive flexibility inherent in artistic tasks—unlike symbolic or geometric challenges with strict specifications, artistic prompts (e.g., "create a cat") allow models to produce plausible representations without exact dimensional requirements. Gemini Pro remains weak at 19.2\%, particularly struggling with vehicles (9.7\%), suggesting its difficulties extend beyond rigid specifications to general spatial understanding.

\subsection{Human Evaluation Analysis}
\label{sec:human_eval_results}

Using the multi-dimensional annotation scheme shown in Figure~\ref{fig:humanAnno}, we analyze model performance across five evaluation axes. While GPT-5 and Claude Sonnet 4.5 achieve similar shape correctness (87.9\% vs.\ 80.4\%), GPT-5 demonstrates higher visual quality scores (5.71 vs.\ 5.01), suggesting better aesthetic judgment in output generation. The strong correlation between Has Object and Shape Correct metrics (difference $<$3\% for most models) indicates that when models produce visible output, they typically generate correct shapes.

Gemini Pro's consistently low performance across all metrics (Has Object: 20.1\%, Shape: 19.5\%, Look: 1.36) points to systematic failure rather than partial success---the model frequently generates code that either fails to execute or produces empty outputs.

\subsection{Ablation: Effect of Additional Working Examples}
\label{sec:ablation}

We investigate whether providing additional working examples in the prompt improves model performance. Our baseline prompt includes 2--3 minimal examples demonstrating API usage, while the extended prompt includes 5+ comprehensive examples covering more API functions and edge cases.

\begin{table}[t]
\centering
\caption{\textbf{Ablation: Effect of additional working examples.} Shape correctness (\%) on artistic composition tasks comparing baseline (2--3 examples) vs.\ extended prompts (5+ examples). $\Delta$ indicates percentage point change.}
\label{tab:ablation_examples}
\resizebox{0.5\textwidth}{!}{%
\begin{tabular}{lccc}
\toprule
\textbf{Model} & \textbf{Baseline} & \textbf{More Ex.} & \textbf{$\Delta$} \\
\midrule
GPT-5 & 97.5 & 96.2 & $-$1.4 \\
GPT-5 Mini & 92.5 & 96.2 & +3.7 \\
GPT-5 Chat & 94.9 & 61.5 & $-$33.4 \\
Claude Sonnet 4.5 & 89.5 & 72.0 & $-$17.5 \\
Claude 3.5 Sonnet & 84.8 & 50.0 & $-$34.8 \\
Gemini Pro & 19.2 & 44.0 & +24.8 \\
\bottomrule
\end{tabular}}
\end{table}

Table~\ref{tab:ablation_examples} reveals a surprising pattern: additional examples help low-performing models but hurt high-performing ones. Gemini Pro shows the largest improvement (+24.8pp), suggesting that its baseline failures stem partly from insufficient API understanding. However, Claude 3.5 Sonnet ($-$34.8pp), GPT-5 Chat ($-$33.4pp), and Claude Sonnet 4.5 ($-$17.5pp) show substantial degradation with more examples, likely due to context dilution or example overfitting. GPT-5 remains stable ($-$1.4pp), suggesting robustness to prompt length variation.

\subsection{Error Analysis}
\label{sec:error_analysis}
To understand the failure modes of different models, we manually analyzed execution logs. As shown in Table~\ref{tab:error_analysis}, we identified four primary error categories with notably different distributions across models.

\begin{table}[t]
\centering
\caption{\textbf{Error type distribution.} Common errors encountered during code execution. Error rate shows percentage of submissions with errors for each model.}
\label{tab:error_analysis}
\resizebox{0.7\textwidth}{!}{%
\begin{tabular}{lcccc}
\toprule
\textbf{Error Type} & \textbf{GPT-5} & \textbf{Claude Son. 4.5} & \textbf{Gemini Pro} & \textbf{Total} \\
\midrule
Invalid API Attribute & 6 & 8 & 15 & 89 \\
Execution Timeout/Crash & 2 & 10 & 1 & 55 \\
Type Unpacking Error & 0 & 0 & 12 & 15 \\
Syntax Error & 0 & 0 & 14 & 15 \\
\midrule
\textbf{Error Rate} & 0.8\% & 1.6\% & 5.6\% & -- \\
\bottomrule
\end{tabular}}
\end{table}

The most frequent failures stem from invalid API usage, accounting for nearly 40\% of all errors. Models frequently hallucinate plausible-sounding but non-existent function names—for instance, calling \texttt{EVoxelBooleanOperation.Intersection} or \texttt{.Subtraction} instead of the correct API names. This pattern suggests models are extrapolating from general programming knowledge rather than carefully following the provided documentation.
The second major category is execution timeouts and rendering crashes, which make up about a quarter of failures. These occur when generated code is syntactically valid but produces computationally infeasible operations, such as infinite loops or degenerate geometric constructions. Interestingly, Claude Sonnet 4.5 is particularly prone to this failure mode—over half of its errors (10 out of 19) fall into this category, while GPT-5 and Gemini Pro rarely encounter it. This suggests Claude generates syntactically correct code that nonetheless misunderstands computational constraints.

The error profiles reveal striking differences in model reliability. GPT-5 achieves the lowest error rate at 0.8\%, with failures almost exclusively in the API hallucination category. In contrast, Gemini Pro exhibits a 5.6\% error rate with failures spanning all categories, including 14 basic syntax errors—suggesting fundamental difficulties with code generation. Claude Sonnet 4.5 falls between these extremes, producing mostly valid syntax but occasionally generating computationally problematic code.

\section{Conclusion}
\label{sec:conclusion}
We introduce \textsc{VoxelCodeBench}, a benchmark for evaluating spatial reasoning through API-driven 3D voxel manipulation. Built on Unreal Engine and Voxel Plugin 2.0, our open-source VoxelCode platform enables real-time rendering of LLM-generated code, providing researchers with extensible infrastructure for studying code-based 3D generation. Our experiments reveal a key finding: voxel-based representations serve as an effective interface for LLMs to externalize their geometric knowledge. Models successfully generate diverse objects by composing primitives through boolean operations, demonstrating that pretrained world knowledge can generalize to 3D structure synthesis without domain-specific training. We release VoxelCode and all evaluation tools to support future research in code-based 3D generation and spatial reasoning.

\clearpage
\section*{Impact Statement}
  This work aims to advance the understanding of spatial reasoning capabilities in large language models through systematic
  benchmarking. We discuss potential societal implications below.

\paragraph{Positive impacts.}
VoxelCodeBench may benefit the research community by providing a standardized evaluation framework for measuring progress
in code-based 3D generation, an area lacking systematic benchmarks. The code-based approach offers interpretability
advantages over end-to-end neural methods, as generated programs expose the reasoning process for analysis and debugging.
Our open-source release of benchmark infrastructure—including tasks, rendering environment, and evaluation tools—lowers
barriers for researchers studying spatial reasoning. In application contexts, improved code generation for 3D manipulation
could facilitate human-AI collaboration in content creation, architectural visualization, and educational tools,
potentially democratizing access to 3D modeling capabilities.

  \paragraph{Potential risks and limitations.}
  We identify several considerations. First, benchmark performance should not be over-interpreted as evidence of general
  spatial intelligence; our evaluation measures specific API-driven tasks that may not transfer to broader 3D reasoning
  scenarios. Second, advances in automated 3D content generation could potentially be misused to create misleading
  visualizations, though the voxel-based outputs in our benchmark are stylized rather than photorealistic, limiting this
  concern. Third, the computational resources required for large-scale model evaluation carry environmental costs, which we
  partially mitigate by focusing on API-based inference rather than training. Finally, while our benchmark evaluates existing
   model capabilities, we do not anticipate that this evaluation-focused work directly enables harmful applications beyond
  those already possible with current systems.

\clearpage
\newpage

\bibliographystyle{assets/plainnat}
\bibliography{main}

\clearpage
\newpage
\beginappendix

\section{Task Examples}

This appendix provides representative examples from each task category in \textsc{VoxelCodeBench}.

\subsection{Symbolic Reasoning Tasks}

\paragraph{Example 1: Primitive Instantiation}
\begin{verbatim}
Task: Create a sphere with radius 5 centered
at position (10, 0, 10).

Expected API calls:
sphere = MakeSphereNode(
    Radius=5.0,
    Transform=FTransform(Location=(10, 0, 10))
)
MakeStampFromNode(sphere, SurfaceType, Layer, Transform)
\end{verbatim}

\paragraph{Example 2: Multiple Primitives}
\begin{verbatim}
Task: Place a torus above a cube, both centered
at the origin.

Expected API calls:
cube = MakeCubeNode(
    Size=FVector(10, 10, 10),
    Roundness=0.0,
    Transform=FTransform(Location=(0, 0, 5))
)
torus = MakeTorusNode(
    MajorRadius=8.0,
    MinorRadius=2.0,
    Transform=FTransform(Location=(0, 0, 15))
)
UnionShapes(Stamp=stamp, Shapes=[cube, torus], Smoothness=0.0)
\end{verbatim}

\subsection{Geometric Construction Tasks}

\paragraph{Example 3: Boolean Composition}
\begin{verbatim}
Task: Create a cube with side length 10, then carve
out a sphere with radius 4 from its center.

Expected approach using boolean subtraction:
cube = MakeCubeNode(
    Size=FVector(10, 10, 10),
    Roundness=0.0,
    Transform=FTransform()
)
sphere = MakeSphereNode(Radius=4.0, Transform=FTransform())
SubtractShapes(
    Stamp=stamp,
    BaseShape=cube,
    ShapesToSubtract=[sphere],
    Smoothness=2.0,
    Transform=FTransform()
)
\end{verbatim}

\paragraph{Example 4: Iterative Construction}
\begin{verbatim}
Task: Create a spiral staircase with 8 steps, each
rotated 45 degrees and elevated by 2 units.

Expected approach involves loops:
steps = []
for i in range(8):
    angle = i * 45
    height = i * 2
    step = MakeCubeNode(
        Size=FVector(20, 50, 5),
        Roundness=0.0,
        Transform=FTransform(
            Location=(0, 0, height),
            Rotation=(0, 0, angle)
        )
    )
    steps.append(step)
UnionShapes(Stamp=stamp, Shapes=steps, Smoothness=0.0)
\end{verbatim}

\paragraph{Example 5: Curved Geometry}
\begin{verbatim}
Task: Create a curved pipe connecting two points
with smooth bends.

Expected approach using Hermite curves:
pipe = MakeHermitePipeNode(
    StartPosition=FVector(0, 0, 0),
    StartVelocity=FVector(100, 0, 0),
    EndPosition=FVector(200, 100, 50),
    EndVelocity=FVector(100, 0, 0),
    PipeOuterRadius=5.0,
    PipeInnerRadius=3.0,
    bClosedPipeEnds=True,
    NumSegments=20,
    Smoothness=10.0,
    Transform=FTransform()
)
\end{verbatim}

\subsection{Artistic Composition Tasks}

\paragraph{Example 6: Thematic Scene}
\begin{verbatim}
Task: Build a small medieval castle with four corner
towers and a central keep.

This open-ended task requires creative problem-solving.
A solution might combine:
- MakeHouseNode for the central keep structure
- MakeCylinderNode for cylindrical towers
- MakeCubeNode for walls and battlements
- MakePyramidNode for conical tower roofs
- UnionShapes to combine all components

The model must determine appropriate dimensions,
positions, and proportions to create a coherent
architectural composition.
\end{verbatim}

\paragraph{Example 7: Natural Forms}
\begin{verbatim}
Task: Create a simple tree with trunk and foliage.

Expected approach combining primitives:
trunk = MakeCylinderNode(
    Radius=3.0, HalfHeight=20.0,
    Transform=FTransform(Location=(0, 0, 20))
)
foliage = MakeSphereNode(
    Radius=15.0,
    Transform=FTransform(Location=(0, 0, 50))
)
UnionShapes(Stamp=stamp, Shapes=[trunk, foliage],
            Smoothness=5.0)
\end{verbatim}

\section{Structured Geometric Detail Generation}
\begin{figure}[h]
\centering
\includegraphics[width=0.9\textwidth]{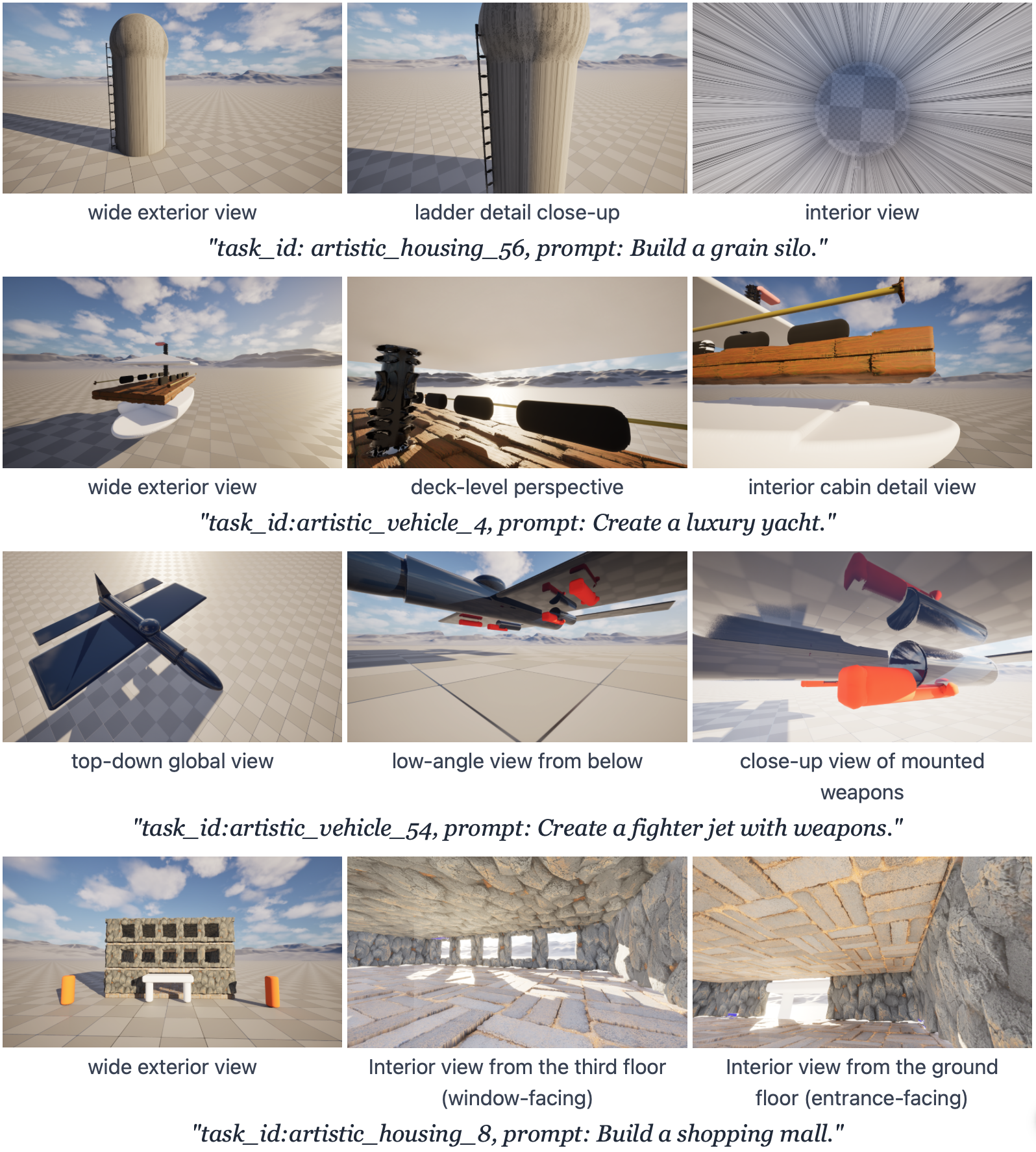}
\caption{\textbf{Structured geometric detail generation.} Code-based generation produces objects with consistent internal
structures (interior views) and fine-grained details (ladders, weapons, floor layouts) that are difficult to achieve with
surface-based neural 3D generation methods.}
\label{fig:structure}
\end{figure}

A key advantage of code-based 3D generation is the ability to produce objects with coherent internal structures. As shown
in Figure~\ref{fig:structure}, our approach generates geometries that can be inspected from multiple viewpoints, including
interior views. For example, the grain silo includes an internal ladder structure, the luxury yacht features a detailed
cabin interior, and the shopping mall has distinct floor layouts visible from inside.
This capability arises because LLM-generated Python code constructs objects through logical composition of geometric
primitives and boolean operations, naturally producing consistent internal geometry. In contrast, diffusion-based 3D
generation methods typically produce surface-only representations optimized for external appearance, making it difficult to
generate meaningful internal structures or fine-grained geometric details such as mounted weapons or architectural
interiors.

\section{Qualitative Analysis of Model Behavior}
\label{sec:qualitative_analysis}

Beyond aggregate metrics, our detailed examination of model outputs reveals nuanced patterns in geometric reasoning and API usage that illuminate both capabilities and limitations of current LLMs.

\paragraph{Geometric concept understanding.}
Models exhibit significant variation in understanding geometric terminology. For instance, only GPT-5 and Claude Sonnet 4.5 correctly interpret ``frustum'' (a truncated cone or pyramid), while other models produce unrelated shapes. Similarly, only Claude Sonnet 4.5 correctly generates a ``prism'' shape---other models confuse it with pyramids or cuboids. For ``capsule'' shapes, we observe diverse misinterpretations: Claude Sonnet 4.5 produces an ellipse, Claude 3 Opus generates a cylinder, while Claude Opus 4 succeeds. These findings suggest that geometric vocabulary understanding varies substantially across model families and does not consistently correlate with overall benchmark performance.

\paragraph{Compositional reasoning for complex shapes.}
When generating shapes not directly available as API primitives (e.g., paraboloids, hyperboloids, heart shapes), models employ different compositional strategies. Some models (GPT-5, Claude Sonnet 4.5) leverage the \texttt{MakeMeshNode} API to construct surfaces from vertex data, while others (Gemini Pro, Claude 3 Opus) attempt to approximate shapes by stacking cylinders or combining primitives. Notably, even simple shapes like hearts prove challenging---models repeatedly fail despite multiple attempts, suggesting that certain geometric forms require spatial reasoning capabilities that current LLMs lack.

\paragraph{Boolean operation challenges.}
Boolean operations (union, intersection, subtraction) present particular difficulty. For tasks requiring cylinder intersections, only GPT-5 produces correct results across multiple test cases. Similarly, for cone-cylinder joins at their bases, only GPT-5 consistently succeeds. This pattern suggests that reasoning about volumetric set operations and spatial relationships between multiple objects remains a significant challenge.

\paragraph{Reproducibility and consistency.}
A striking observation is the inconsistency of model performance across similar tasks. The same model may succeed on one instance of a shape (e.g., crescent, dome, pentagonal prism) but fail on another instance with slightly different parameters or phrasing. For example:
\begin{itemize}[leftmargin=*,noitemsep,topsep=2pt]
\item GPT-5 correctly generates a crescent in one task but fails on a similar crescent task
\item GPT-5 Chat succeeds on one pentagonal prism task but fails on another
\item GPT-5 generates correct domes in some cases but incorrect ones in others
\end{itemize}
This inconsistency suggests that model performance is sensitive to prompt variations and may not reflect robust geometric understanding.

\paragraph{API hallucination patterns.}
Models frequently hallucinate plausible-sounding but non-existent API functions. Common hallucinations include:
\begin{itemize}[leftmargin=*,noitemsep,topsep=2pt]
\item \texttt{EVoxelBooleanOperation.Difference} / \texttt{.Subtraction} / \texttt{.Intersection} (incorrect enum names)
\item \texttt{MakePrismNode}, \texttt{MakeConeNode} (non-existent functions)
\item \texttt{SetStampWithSurface}, \texttt{log\_error} (invalid method calls)
\end{itemize}
These hallucinations suggest models rely on general programming knowledge and naming conventions rather than faithfully following the provided API documentation. This behavior is particularly pronounced in Gemini Pro, which exhibits the highest rate of API-related errors.

\paragraph{Implications.}
These observations highlight that current LLMs, while capable of impressive code generation, exhibit brittle geometric reasoning. Performance on individual tasks does not reliably predict performance on similar tasks, and models frequently substitute plausible guesses for careful API adherence. Future work on improving spatial reasoning should address both geometric concept grounding and robust API comprehension from documentation.

\newpage

\end{document}